# Maximum Margin Bayesian Networks

**Yuhong Guo**
Department of Computing Science
University of Alberta
yuhong@cs.ualberta.ca

**Dana Wilkinson**
School of Computer Science
University of Waterloo
d3wilkin@cs.uwaterloo.ca

**Dale Schuurmans**
Department of Computing Science
University of Alberta
dale@cs.ualberta.ca

## Abstract

We consider the problem of learning Bayesian network classifiers that maximize the margin over a set of classification variables. We find that this problem is harder for Bayesian networks than for undirected graphical models like maximum margin Markov networks. The main difficulty is that the parameters in a Bayesian network must satisfy additional normalization constraints that an undirected graphical model need not respect. These additional constraints complicate the optimization task. Nevertheless, we derive an effective training algorithm that solves the maximum margin training problem for a range of Bayesian network topologies, and converges to an approximate solution for arbitrary network topologies. Experimental results show that the method can demonstrate improved generalization performance over Markov networks when the directed graphical structure encodes relevant knowledge. In practice, the training technique allows one to combine prior knowledge expressed as a directed (causal) model with state of the art discriminative learning methods.

## 1 Introduction

When training probability models for classification tasks it is often recommended that the model parameters be optimized under a discriminative training criterion such as conditional likelihood (Friedman et al., 1997; Lafferty et al., 2001; Lafferty et al., 2004). However, general Bayesian network classifiers have rarely, if ever, been trained to maximize the margin—arguably the most discriminative criterion available. Recently, it has been observed that undirected graphical models can be efficiently trained to maximize the margin, even simultaneously, over a set of classification variables (Taskar et al., 2003; Taskar et al., 2004; Altun et al., 2003; Tsochantaridis et al., 2004). However, these training algorithms have adopted the Euclidean normalization constraint of support vector machines (SVMs), which can be accommodated in their frameworks because they rely on an undirected graphical model representation which allows a single arbitrary normalization.

In this paper we consider applying the maximum margin methodology to Bayesian networks; that is, directed graphical models. Unlike Markov network models, Bayesian networks require that additional normalization constraints be satisfied; namely that the local clique potentials represent conditional probability distributions. These constraints are very different from the standard Euclidean normalization constraints of SVMs. Nevertheless, they do not preclude the possibility of learning large margin classifiers. Our goal is simply to exploit the benefits of learning maximum margin classifiers, while still being able to represent the learned classifier as a Bayesian network.

There are several motivations for attempting to maintain a Bayesian network representation. First, the classification model being learned could be a fragment of a much larger probabilistic causal model. In this case, maintaining a Bayesian network representation could allow one to integrate the learned model with a pre-existing background model without additional effort. Second, the normalization constraints asserted by a directed graphical structure capture nonparametric *causal* knowledge about the domain. Therefore, respecting these constraints allows one to exploit the advantages of Bayesian networks for capturing intuitive causal structure. Note that removing the normalization constraints would turn the Bayesian network into a Markov network, and this would necessarily remove the causal knowledge that was originally encoded by the local normalization constraints.

To understand both the prospects and limitations of learning maximum margin Bayesian network classifiers we proceed as follows. First, after preliminary definitions in Section 2, we investigate the notion of *classification margin* for Bayesian network classifiers in Section 3, and relate this to the common conditional likelihood criterion of graphical models. We then present a convex relaxation in Section 4 that can be used to derive an effective training algorithm (Section 5). The algorithm solves a wide range of problems exactly and otherwise provides an effective heuristic for finding approximate solutions (Section 6). In Section 7 we then present experimental results which show that the causal information in Bayesian networks can yield effective generalization performance when the directed graphical structure captures relevant causal knowledge. Finally, we extend the approach to multivariate classification in Section 8 and present further experimental results in Section 9.

## 2 Bayesian networks

We assume we are given a Bayesian network which is defined by a directed acyclic graph over variables $X_1, ..., X_n$, where the probability of a configuration $\mathbf{x}$ is given by

$$\begin{aligned} P(\mathbf{x}|\boldsymbol{\theta}) &= \prod_{j=1}^{n} P(x_j|\mathbf{x}_{\pi(j)}) \\ &= \exp\left(\sum_{ja\mathbf{b}} 1_{(\mathbf{x}_j = a\mathbf{b})} \ln \theta_{ja\mathbf{b}}\right) \end{aligned} \quad (1)$$

Here $\boldsymbol{\theta}$ denotes the parameters of the model; $j$ ranges over conditional probability tables (CPTs), one for each variable $X_j$; $1_{(\cdot)}$ denotes the indicator function; $\mathbf{x}_j$ denotes the local subconfiguration of $\mathbf{x}$ on $(x_j, \mathbf{x}_{\pi(j)})$; $a$ denotes the set of values for child variable $x_j$; and $\mathbf{b}$ denotes the set of configurations for $x_j$'s parents $\mathbf{x}_{\pi(j)}$. The form (1) shows how Bayesian networks are a form of exponential model

$$P(\mathbf{x}|\mathbf{w}) = \exp\left(\boldsymbol{\phi}(\mathbf{x})^\top \mathbf{w}\right) \quad (2)$$

using the substitution $w_{ja\mathbf{b}} = \ln \theta_{ja\mathbf{b}}$, where $\boldsymbol{\phi}(\mathbf{x})$ denotes the feature vector $(...1_{(\mathbf{x}_j=a\mathbf{b})}...)^\top$ over $j, a, \mathbf{b}$. The key aspect of the exponential form is that it expresses $p(\mathbf{x}|\mathbf{w})$ as a convex function of the parameters $\mathbf{w}$, which would seem to suggest convenient optimization problems. However, Bayesian networks also require the imposition of additional normalization constraints over each variable

$$\sum_a e^{w_{ja\mathbf{b}}} = 1 \text{ for all } j, \mathbf{b} \quad (3)$$

Unfortunately, these constraints are nonlinear, even though the log-objective is linear in $\mathbf{w}$. Removing these constraints greatly reduces the computational challenge of training, but also removes the causal interpretability of the model. Therefore, our goal in this paper is to maintain the Bayesian network constraints while investigating the consequences.

## 3 Discriminative training criteria

We initially assume there is a single classification variable $Y$ taking on values $y \in \{1, .., K\}$. (We will extend this to multiple classification variables in Section 8 below.) To make predictions, one usually considers the maximum conditional probability prediction $\arg\max_y P(y|\mathbf{x})$. Note that for graphical models the conditional probability depends only on variables that share some common function (CPT) with $Y$ (the Markov blanket of $Y$), and therefore we will restrict attention to this set of variables henceforth.

We are interested in learning the parameters for a Bayesian network classifier given training data of the form $(\mathbf{x}^1 y^1), ..., (\mathbf{x}^T y^T)$. Two standard training criteria to maximize during training are the joint loglikelihood and the conditional loglikelihood, given respectively by

$$\begin{aligned} \log L(\boldsymbol{\theta}) &= \sum_{i=1}^{T} \log P(y^i|\mathbf{x}^i, \boldsymbol{\theta}) + \log P(\mathbf{x}^i|\boldsymbol{\theta}) \quad (4) \\ \log CL(\boldsymbol{\theta}) &= \sum_{i=1}^{T} \log P(y^i|\mathbf{x}^i, \boldsymbol{\theta}) \quad (5) \end{aligned}$$

Much of the literature suggests that the latter objective is better suited for classification (Lafferty et al., 2001; Friedman et al., 1997), although conditions have been identified where the former is advantageous (Ng & Jordan, 2001; Raina et al., 2003).

In this paper we investigate an alternative criterion based on the large margin criteria of SVMs. In particular, we adopt the multiclass margin definition of (Crammer & Singer, 2001). In our context, this objective can be cast maximizing the minimum conditional likelihood ratio (MCLR)

$$\begin{aligned} MCLR(\boldsymbol{\theta}) &= \min_{i=1}^{T} \min_{y \neq y^i} \frac{P(y^i|\mathbf{x}^i, \boldsymbol{\theta})}{P(y|\mathbf{x}^i, \boldsymbol{\theta})} \\ \log MCLR(\boldsymbol{\theta}) &= \min_{i=1}^{T} \min_{y \neq y^i} \begin{array}{l} \log P(\mathbf{x}^i, y^i|\boldsymbol{\theta}) \\ - \log P(\mathbf{x}^i, y|\boldsymbol{\theta}) \end{array} \quad (6) \end{aligned}$$

Thus our goal is to find a set of parameters $\boldsymbol{\theta}$ that *maximizes* the minimum margin between the target classification label against the best alternative under the probability model. (We introduce slack variables to obtain a soft margin version of the criterion below.)

To see the connection to SVMs more clearly, note that one can substitute the exponential form of $P(\mathbf{x}, y|\mathbf{w})$ into the MCLR objective, to obtain

$$\begin{aligned} \log MCLR(\mathbf{w}) &= \min_{i=1}^{T} \min_{y \neq y^i} \left[\boldsymbol{\phi}(\mathbf{x}^i, y^i) - \boldsymbol{\phi}(\mathbf{x}^i, y)\right]^\top \mathbf{w} \\ &= \min_{i=1}^{T} \min_{y \neq y^i} \boldsymbol{\Delta}(i, y) \mathbf{w} \quad (7) \end{aligned}$$

where $\boldsymbol{\Delta}(i, y) = [\boldsymbol{\phi}(\mathbf{x}^i, y^i) - \boldsymbol{\phi}(\mathbf{x}^i, y)]^\top$. Here the row vector $\boldsymbol{\Delta}(i, y)$ plays the role of the feature vector for training example $i$ and class label $y$, and therefore we can write the entire set of feature vectors as a matrix $\boldsymbol{\Delta}$ of size $(TK) \times$ (number of features).

Thus, starting with the training objective (6), through a change of parameters, we are led to a training problem that can be cast as a conventional maximum margin problem

$$\max_{\mathbf{w},\gamma} \gamma \quad \text{subject to} \quad \Delta \mathbf{w} \geq \gamma \boldsymbol{\delta}, \quad \|\mathbf{w}\| \leq 1 \quad (8)$$

where $\delta_{(i,y)} = 1_{(y \neq y^i)}$. Notice that here we have added the normalization constraint $\|\mathbf{w}\| \leq 1$. Obviously some form of normalization is necessary to avoid making $\Delta \mathbf{w}$ large in a trivial manner just by making $\mathbf{w}$ large. Euclidean normalization happens to yield a weight vector that maximizes the Euclidean margin (Schoelkopf & Smola, 2002). The resulting constrained optimization problem corresponds to the standard version of multiclass SVMs proposed in (Crammer & Singer, 2001) (ignoring slacks), here expressed over features determined by the Bayesian network.

In fact, this is the connection between probabilistic and large margin classifiers that is one of the main observations of (Taskar et al., 2003; Altun et al., 2003), who then proceed to use standard SVM training criteria over these features. Note however that the solution weight vector for (8) cannot be substituted into the Bayesian network representation, because it will not satisfy the proper normalization constraints (3). The previous techniques of (Taskar et al., 2003; Altun et al., 2003) were able to proceed by using an *undirected* graphical model which could accommodate unnormalized weights in the potential function. However, for Bayesian networks this is not sufficient, and there is usually no way to represent the same classifier in the original Bayesian network structure.

The alternative approach we consider, therefore, is to maximize the same objective, but subject to constraints that preserve representability as a Bayesian network

$$\max_{\mathbf{w},\gamma} \gamma \quad \text{subject to} \quad \Delta \mathbf{w} \geq \gamma \boldsymbol{\delta}, \quad \sum_a e^{w_{ja\mathbf{b}}} = 1 \; \forall j, \mathbf{b} \quad (9)$$

Unfortunately, these natural constraints on $\mathbf{w}$ are nonlinear and this yields a difficult optimization problem. Attempts to reformulate the problem according to standard transformations also fail. For example, the probability function (1) is neither concave nor convex in the parameters $\theta$, even though the equality constraints (3) are linear in $\theta$. The standard trick of removing the normalization constraints via the transformation $\theta_{ja\mathbf{b}} = e^{\omega_{ja\mathbf{b}}} / \sum_a e^{\omega_{ja\mathbf{b}}}$ also does not work in this case, since it creates terms of the form $\sum_{ja\mathbf{b}} \Delta_{(j,a,\mathbf{b})}(i,y) [\omega_{ja\mathbf{b}} - \log \sum_a e^{\omega_{ja\mathbf{b}}}]$ which are neither convex nor concave in $\boldsymbol{\omega}$. Thus, if we hope to solve the maximum margin Bayesian network training problem exactly, we require a more subtle approach.

## 4 Convex relaxation

Although solving for the maximum margin Bayesian network parameters appears to be a hard problem, we can derive a practical training algorithm that still solves the problem for a wide range of graph topologies, and otherwise provides a useful foundation for approaches that seek local maxima. The main idea is to try to exploit convexity in the problem as much as possible, and identify situations where the solutions to a convex subproblem can be maintained.

First note that the objective in (9) is a linear function of $\mathbf{w}$. Unfortunately, the normalization constraints are nonlinear equalities on $\mathbf{w}$, which eliminates the convexity of the problem. However, our basic observation is that the problem can be made convex simply by relaxing these equality constraints to inequality constraints, thus yielding a simple relaxation

$$\max_{\mathbf{w},\gamma} \gamma \quad \text{subject to} \quad \Delta \mathbf{w} \geq \gamma \boldsymbol{\delta}, \quad \sum_a e^{w_{ja\mathbf{b}}} \leq 1 \; \forall j, \mathbf{b} \quad (10)$$

The solution to this problem will of course be subnormalized. The key fact about the relaxed problem (10) however, is that it is convex in $\mathbf{w}$ and this will permit effective algorithmic approaches (Boyd & Vandenberghe, 2004).[1]

It is interesting to compare the two convex optimization problems (8) and (10), which correspond to maximum margin Markov networks and Bayesian networks respectively. These problems have identical objectives and margin constraints on $\mathbf{w}$, but differ only in the normalization constraints—one global constraint for Markov networks versus multiple local constraints for Bayesian networks. The solutions to the two problems will obviously be different. Intuitively, the Bayesian network constraints might regularize the weights more comprehensively in the sense that each local CPT is constrained to have identical maximum influence, whereas a Markov network could concentrate its weight in a single local function.

### 4.1 Slack variables

Before tackling real problems we need to introduce slack variables, since it is obviously not practical to use a hard margin formulation on real data. To this end, we consider the standard soft margin formulation of SVMs

$$\min_{\mathbf{w},\boldsymbol{\xi}} \frac{1}{2}\|\mathbf{w}\|^2 + C\boldsymbol{\xi}^\top \mathbf{e} \quad \text{subject to} \quad \Delta \mathbf{w} \geq \boldsymbol{\delta} - S\boldsymbol{\xi} \quad (11)$$

where $\boldsymbol{\xi}$ are the slack variables; $\mathbf{e}$ denotes the vector of all 1 entries; $S$ is a $TK \times T$ sparse matrix with nonzero entries $S((i,y),i) = 1$ (which enforces the constraint that $\boldsymbol{\Delta}(i,y)\mathbf{w} \geq \gamma \delta_{(i,y)} - \xi_i$ for all $i, y$); and $C$ is a parameter that controls the slack effect (Crammer & Singer, 2001).[2] For our purposes, we need to state this objective explicitly in terms of the margin $\gamma$. It can be shown that (11) is equivalent to (Lanckriet et al., 2004)

$$\min_{\mathbf{w},\gamma,\boldsymbol{\xi}} \frac{1}{2\gamma^2} + C\boldsymbol{\xi}^\top \mathbf{e} \quad \text{subject to} \quad \Delta \mathbf{w} \geq \gamma(\boldsymbol{\delta} - S\boldsymbol{\xi}),$$
$$\|\mathbf{w}\| \leq 1 \quad (12)$$

---

[1] Note that the inequality form of the norm constraints in (8) and (9) is not vacuous: In either case, reducing the magnitude of the weights only has the effect of reducing the inner products in the margin constraints ($\Delta \mathbf{w}$), which can only yield a smaller margin $\gamma$. The maximization objective naturally forces the weight magnitudes overall to become large as possible, subject to the normalization constraints.

Thus, by replacing the Euclidean constraint with the Bayesian network subnormalization constraint, we obtain

$$\min_{\mathbf{w},\gamma,\boldsymbol{\xi}} \frac{1}{2\gamma^2} + C\boldsymbol{\xi}^\top \mathbf{e} \quad \text{subject to} \quad \Delta \mathbf{w} \geq \gamma(\boldsymbol{\delta} - S\boldsymbol{\xi}),$$
$$\sum_a e^{w_{ja\mathbf{b}}} \leq 1 \;\; \forall j, \mathbf{b} \quad (13)$$

The two problems, (12) and (13), specify the soft margin formulations of maximum margin Markov and Bayesian networks respectively. Unfortunately, neither formulation is convex because the quadratic term $\gamma(\boldsymbol{\delta} - S\boldsymbol{\xi})$ is nonconvex in the optimization variables $\gamma$ and $\boldsymbol{\xi}$ (Boyd & Vandenberghe, 2004). For Markov networks one can simply convert (12) back to (11) and thus convexify the problem.[3] For Bayesian networks we can instead solve the following problem with alternative slack variables $\boldsymbol{\epsilon}$ and parameter $B$

$$\min_{\mathbf{w},\gamma,\boldsymbol{\epsilon}} \frac{1}{2\gamma^2} + B\boldsymbol{\epsilon}^\top \mathbf{e} \quad \text{subject to} \quad \Delta \mathbf{w} \geq \gamma\boldsymbol{\delta} - S\boldsymbol{\epsilon},$$
$$\gamma \geq 0, \;\; \sum_a e^{w_{ja\mathbf{b}}} \leq 1 \;\; \forall j, \mathbf{b} \quad (14)$$

Provided $\gamma \geq 0$—which we henceforth assume—the problem formulation (14) is convex and is equivalent to (13). Thus, this formulation yields a convex version of the soft margin training problem for Bayesian networks.

**Proposition 1** *Assuming $\gamma \geq 0$, $(\mathbf{w}, \gamma, \boldsymbol{\xi})$ is an optimal point for $C$ in (13) if and only if $(\mathbf{w}, \gamma, \boldsymbol{\epsilon})$ is an optimal point for $B$ in (14) with $\boldsymbol{\epsilon} = \gamma\boldsymbol{\xi}$ and $B = C/\gamma$.*

So if one chooses an optimal regularization parameter $B$ for (14), then the optimal solution $(\mathbf{w}, \gamma)$ will be preserved, while the slacks $\boldsymbol{\xi}$ can be recovered by $\boldsymbol{\xi} = \boldsymbol{\epsilon}/\gamma$.

We now proceed to develop algorithmic approaches for solving the convex training problem (14), with the goal of ultimately comparing maximum margin Markov networks trained under (11) versus Bayesian networks trained under (14).

## 5 A training algorithm

To solve (14) first consider the Lagrangian

$$L(\mathbf{w}, \gamma, \boldsymbol{\epsilon}, \boldsymbol{\eta}, \boldsymbol{\lambda}, \nu)$$

---

[2]Note that $\boldsymbol{\xi} \geq 0$ is already implied, because $\boldsymbol{\Delta}(i, y^i) = \mathbf{0}$ and $\delta(i, y^i) = 0$ for all $i$.

[3]It is of course no surprise that optimization problems can be converted between convex and nonconvex formulations without affecting the location of the optimal solution.

$$= 1/(2\gamma^2) + B\boldsymbol{\epsilon}^\top \mathbf{e} + \boldsymbol{\eta}^\top(\gamma\boldsymbol{\delta} - S\boldsymbol{\epsilon} - \Delta\mathbf{w})$$
$$+ \sum_{j,\mathbf{b}} \lambda_{j\mathbf{b}} \left( \sum_a e^{w_{ja\mathbf{b}}} - 1 \right) - \nu\gamma \quad (15)$$

The saddle point condition gives us an equivalent problem to (14)

$$\min_{\mathbf{w},\gamma,\boldsymbol{\epsilon}} \max_{\boldsymbol{\eta},\boldsymbol{\lambda},\nu} L(\mathbf{w},\gamma,\boldsymbol{\epsilon},\boldsymbol{\eta},\boldsymbol{\lambda},\nu) \text{ subject to } \boldsymbol{\eta},\boldsymbol{\lambda},\nu \geq 0 \quad (16)$$

Unfortunately, this Lagrangian is not nearly as convenient as the one for the SVM formulation (11), and a closed form solution for the dual is not readily obtainable in this case. For example, one cannot easily eliminate the primal variables from this problem: taking the partial derivative with respect to $w_{ja\mathbf{b}}$ yields

$$\frac{\partial L}{\partial w_{ja\mathbf{b}}} = \lambda_{j\mathbf{b}} e^{w_{ja\mathbf{b}}} - \boldsymbol{\eta}^\top \boldsymbol{\Delta}_{ja\mathbf{b}} \quad (17)$$

where $\boldsymbol{\Delta}_{ja\mathbf{b}}$ denotes the $ja\mathbf{b}$ column of $\boldsymbol{\Delta}$. The difficulty with (17) is that one cannot set this derivative to zero because $\boldsymbol{\eta}^\top \boldsymbol{\Delta}_{ja\mathbf{b}}$ can be negative ($\boldsymbol{\Delta}$ has negative entries). Nevertheless, the problem remains convex.

Rather than use a Lagrangian approach to solve this problem, we instead consider a standard barrier approach. In fact, barrier methods are among the most effective techniques for solving convex constrained optimization problems (Boyd & Vandenberghe, 2004; Vanderbei, 1996). In this approach one simply replaces the constraints with log barrier functions.

$$\min_{\mathbf{w},\gamma,\boldsymbol{\xi}} \frac{1}{2\gamma^2} + B\boldsymbol{\epsilon}^\top \mathbf{e}$$
$$- \mu \sum_{(i,y)} \log \left( \boldsymbol{\Delta}(i,y)\mathbf{w} - \gamma\delta_{(i,y)} + \epsilon_i \right)$$
$$- \mu \sum_{j,\mathbf{b}} \log \left( 1 - \sum_a e^{w_{ja\mathbf{b}}} \right)$$
$$- \mu \log(\gamma) \quad (18)$$

The parameter $\mu$ is initially set to a reasonable value to ensure numerical stability, and then successively reduced to sharpen the barriers. In general, it can be shown that for convex inequality constraints, the resulting unconstrained objective (18) is also convex, while the solution to (18) converges to (14) as $\mu \to 0$ (Boyd & Vandenberghe, 2004). In the standard path following technique, an optimal solution $(\mathbf{w}, \gamma, \boldsymbol{\epsilon})$ is obtained for the current value of $\mu$ (usually using a second order method to ensure fast convergence), after which $\mu$ is decreased, until a small value of $\mu$ is reached.

In our case, for the inner optimization loop, we implemented a Newton descent based on computing the gradient and Hessian of (18) with respect to $(\mathbf{w}, \gamma, \boldsymbol{\epsilon})$. We found that 7 outer iterations, $\mu^{(k+1)} = \mu^{(k)}/10$, $\mu^{(1)} = 1$, and fewer than 20 inner Newton iterations were required to obtain accurate solutions. In principle, the runtime of a barrier iteration method is not dramatically slower than solving a quadratic program (Boyd & Vandenberghe, 2004). However, our Matlab implementation is currently an order of magnitude slower than the quadratic program solver we used (CPLEX). Our largest runtimes in the experiments below are a few minutes, versus a few seconds for CPLEX.

## 6 Exact case: Locally moralized graphs

Before presenting experiments, we first consider when the solutions to the relaxed problem (14) correspond to the solutions to the exact problem; i.e., satisfying (3). Our main concern is that the solutions obtained to (14) may not be representable in a Bayesian network because the parameters **w** are subnormalized, not normalized. This leaves us with the question of determining when these subnormalized solutions can be converted into properly normalized Bayesian networks obeying the correct equality constraints (3).

It turns out that a wide range of network topologies admit a simple procedure for renormalizing the local functions so that they become proper CPTs, without affecting the conditional probability of $y$ given $\mathbf{x}$. In fact, this observation has been previously made by (Wettig et al., 2002; Wettig et al., 2003). We present a simpler view here. In fact, it is easy to characterize when an unnormalized Bayesian network classifier can be renormalized to preserve $P(y|\mathbf{x})$.

**Proposition 2** *An unnormalized directed graphical model, defined by the Markov blanket of $y$, can be renormalized to preserve the decision function $P(y|\mathbf{x})$ if and only if the parents of all children of $y$ are moralized.*

The intuition behind this result is fairly straightforward. Consider an unnormalized local function $f(x, \mathbf{z})$ in a Bayesian network structure and assume we want to normalize it over $x$. Such a function can always be multiplied by a factor $\rho_{\mathbf{z}}$ for each $\mathbf{z}$, as long as there is another local function $f(\mathbf{z}, \mathbf{q})$ that can be divided by the same factor. (That is, a local function that contains *all* the parents $\mathbf{z}$ of $x$.) That is, if an accompanying $f(\mathbf{z}, \mathbf{q})$ always exists, as in Figure 1, we can always renormalize $f(x, \mathbf{z})$. Since the functions and variables follow an acyclic ordering in a Bayesian network, child variables can be sequentially renormalized bottom up without affecting previous normalizations. Finally, the factor containing the $y$ variable can be renormalized to preserve $P(y|\mathbf{x})$.

This renormalization strategy only fails if, at any stage, the parent variable set $\mathbf{z}$ is not contained in a single local function, but is instead split between separate local functions, as in Figure 2. In this case, there would be no way to coordinate the compensation for $\rho_{\mathbf{z}}$ (without adding a new local function over $\mathbf{z}$). Thus, in the end, we are left with an intuitive sufficient condition for when a Bayesian network can be renormalized: Any graph can be normalized without affecting $P(y|\mathbf{x})$ if the child variables can be eliminated without adding any new edges. In these cases, we can recover a normalized model without affecting the optimality of the solution to (14), and thus we obtain a global maximum of (7) with respect to (3).

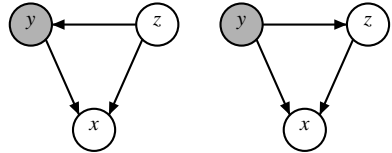

Figure 1: Illustration of renormalizable graphs

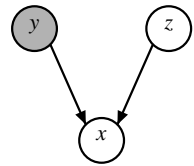

Figure 2: A graph that cannot be renormalized

**Corollary 1** *For a directed graphical model satisfying Proposition 2, (14) is equivalent to satisfying (3) in addition.*

Note that the renormalization procedure can be applied to any set of parameters defining the decision rule $P(y|\mathbf{x})$ in such a network structure, even if the parameters were produced by a Markov network training procedure. However, this does not imply that the resulting model $P(y|\mathbf{x})$ is optimal under the Bayesian network criterion (14).

## 7 Experimental results

To evaluate the utility of learning maximum margin Bayesian networks, we conducted some simple experiments on both real and synthetic data sets. In the synthetic experiments, we fixed a Bayesian network structure and parameters, and used it to generate training and test data. The goal of the synthetic experiments is to run a controlled comparison of maximum margin Bayesian versus Markov networks, to determine the effects of having a correct Bayesian network structure. We experimented with several network topologies and parameterizations, and compared maximum margin Bayesian networks (M$^2$BN) trained according to (14) against maximum margin Markov networks (M$^3$N) trained according to (11), and also against maximum conditional likelihood (MCL). The results are for 100 repetitions of the training sample. For each method, on each model, the regularization parameters, $B$ and $C$ respectively, were optimized on one train/test split and then fixed for the duration of the experiment.

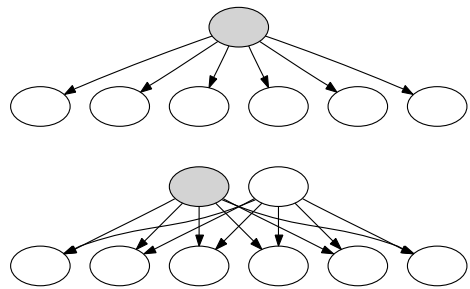

Figure 3: Two Bayesian network models. The classification variable $y$ is shaded.

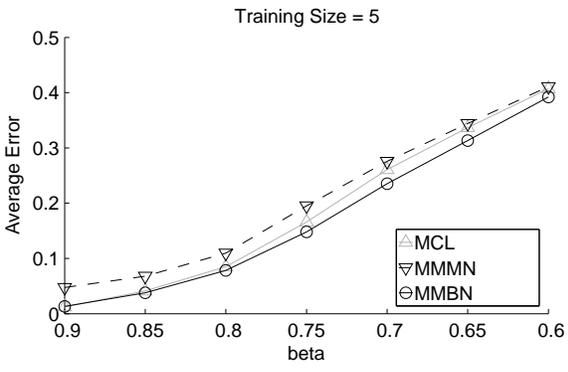
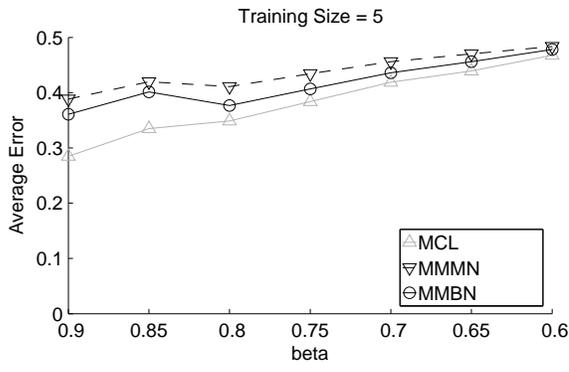
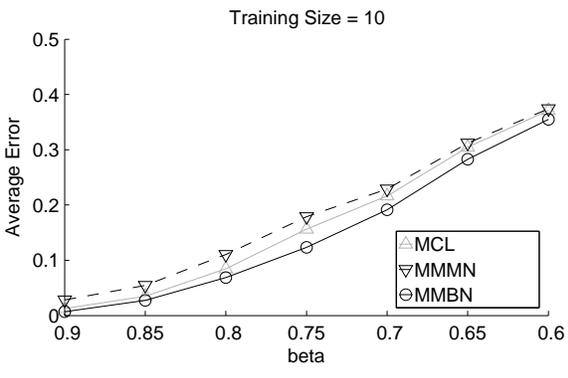
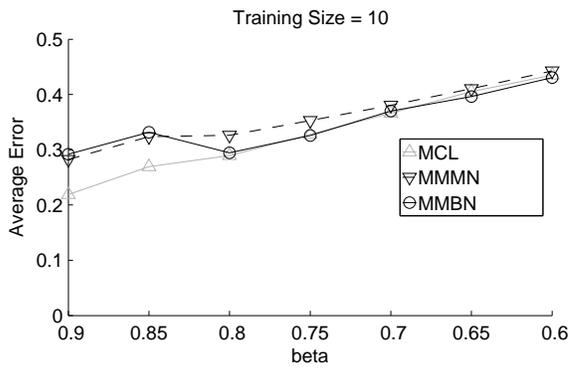
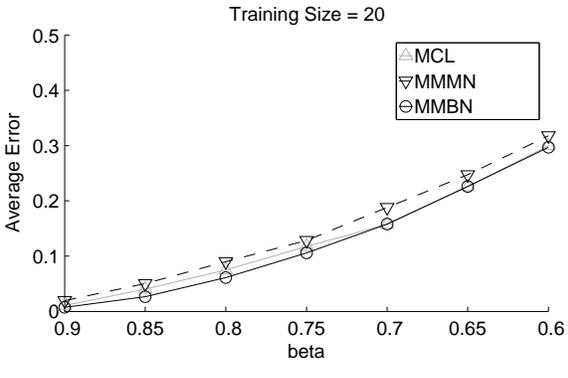
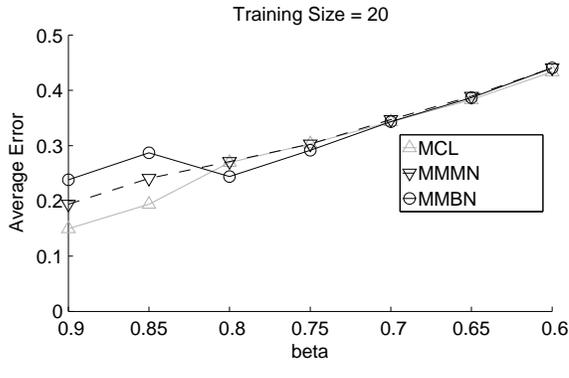
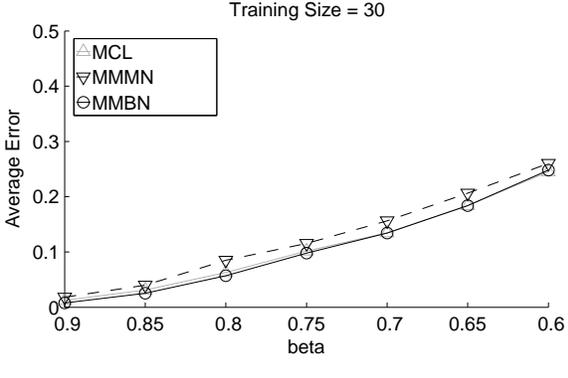
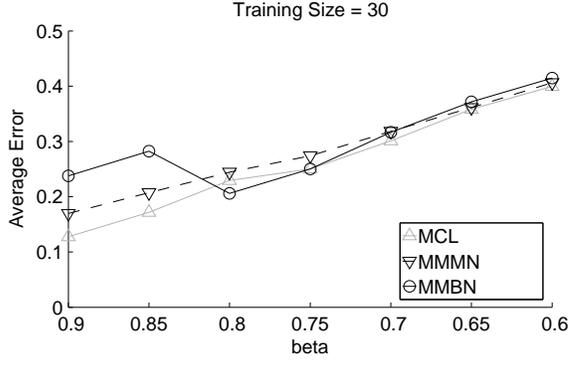

Figure 4: Average error results for Figure 3 (top).

Figure 5: Average error results for Figure 3 (bottom).

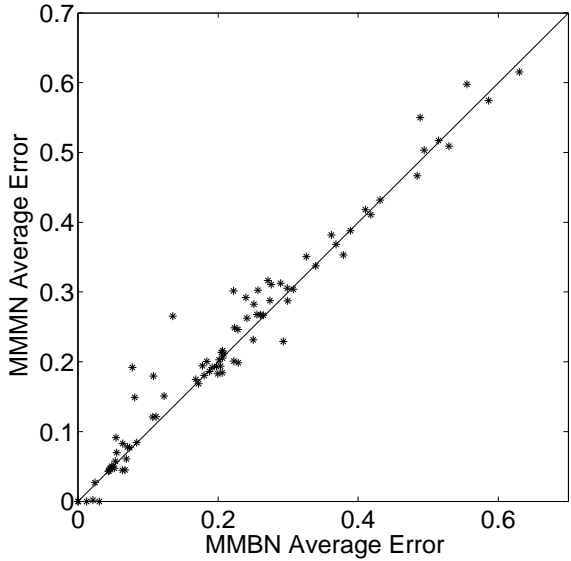

Figure 6: Average error comparison between M$^2$BN and M$^3$N on UCI data sets.

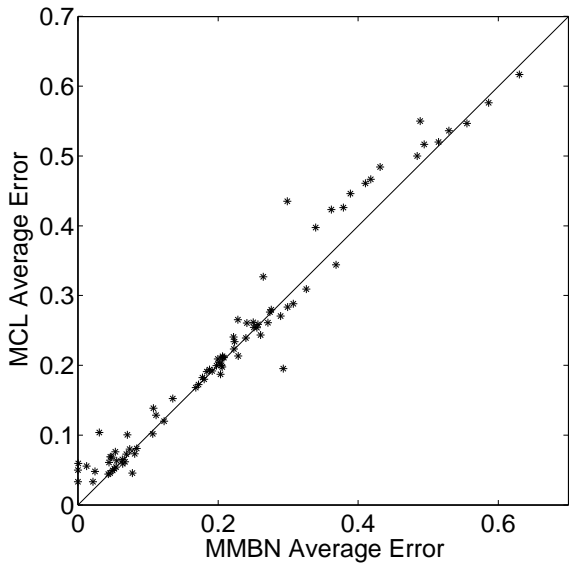

Figure 7: Average error comparison between M$^2$BN and MCL on UCI data sets.

The first synthetic experiments were conducted on the networks shown in Figure 3. Here we fixed a network structure and then defined the generative model by selecting parameters from a skewed distribution. We used a parameter $\beta$ to control the skewness of the conditional distributions of each child, where a value of $\beta = 1$ makes each child a deterministic function of the parents, and $\beta = 0.5$ gives each child a uniform distribution, rendering them effectively independent of their parents. Figure 4 shows a the comparison of the three strategies, M$^2$BN, M$^3$N, and MCL for the first network topology shown in Figure 3. This network topology satisfies the conditions of Proposition 2, and therefore M$^2$BN computes a globally optimal solution in this case. Here we see that for a range of generative models defined by $\beta$ and several training sample sizes, M$^2$BN demonstrates a systematic advantage over both M$^3$N and MCL, although MCL is clearly stronger than M$^3$N in this case. Figure 5 shows the same comparison using the second network topology from Figure 3. This network no longer satisfies the conditions of Proposition 2, and therefore our training algorithm is no longer guaranteed to produce an optimal normalized solution (only an optimal subnormalized solution). Nevertheless, we see that M$^2$BN holds a slight advantage over M$^3$N in this case. MCL proves to be slightly better here. M$^2$BN appears to have an advantage in cases where it is exact (Proposition 2), but the advantage is diminished in the subnormalized case.

For a more realistic comparison, we also experimented with real data from the UCI repository. Specifically, we used 17 data sets: australian*, breast, chess, cleve, diabetes*, flare*, glass, glass2*, heart, hepatitis*, iris*, lymphography, mofn*, pima*, vehicle*, vote, and waveform. In each case, we formulated a Bayesian network topology that was intended to capture the causal structure of the domain, but in this case had no guarantee that the presumed structure was correct. The network structures we used were automatically generated using the "PowerConstructor" technique discussed in (Cheng & Greiner, 1999). These networks are much larger and cannot be easily visualized here. Nevertheless, in 9 of the 17 cases the network topologies satisfied the conditions of Proposition 2 (marked * above), and 8 of the 17 data sets did not. For each data set we considered 5 different training sample sizes, $t = 10, 20, 30, 40, 50$. For each $t$ we sampled 5 different training sets (disjoint where possible), tested on the remainder, and report average results. Interestingly, Figure 6 shows that M$^2$BN obtains an overall advantage over M$^3$N. Perhaps surprisingly, M$^2$BN also shows a slight overall advantage over MCL on these data sets; see Figure 7.

## 8 Multivariate case

Finally, we consider the key extension of (Taskar et al., 2003; Altun et al., 2003; Collins, 2002) to multivariate classification. In this setting, we observe training data $(\mathbf{x}^1, \mathbf{y}^1), ..., (\mathbf{x}^T, \mathbf{y}^T)$ as before, but now the targets $\mathbf{y}^i$ are *vectors* of correlated classifications. The first main issue is to adapt the training criterion (8) to the multivariate prediction case. Following (Taskar et al., 2003) we scale the margin between a target class vector $\mathbf{y}^i$ and an alternative vector $\mathbf{y}$ proportional to the number of misclassifications. That is, we simply set

$$\delta_{(i,\mathbf{y})} = \sum_k 1_{(y_k^i \neq y_k)} \quad (19)$$

for multivariate predictions. This immediately yields multivariate versions of the training problems (11) and (14).

The primary difficulty in dealing with the multivariate form of these problems is coping with the exponential number of constraints in $\mathbf{\Delta w} \geq \gamma \boldsymbol{\delta} - S\boldsymbol{\epsilon}$. That is, one now has to assert $\mathbf{\Delta}(i, \mathbf{y})\mathbf{w} \geq \gamma \boldsymbol{\delta}_{(i,\mathbf{y})} - \epsilon_i$ for all training examples $i$, over all possible label vectors $\mathbf{y}$. Such a constraint set is too large to handle explicitly, and an approach must be developed for handling them implicitly.

One of the key results in (Taskar et al., 2003) is showing that, for maximum margin Markov networks (11), the constrained optimization problem can be factored and re-expressed in terms of "marginal" Lagrange multipliers $\alpha_{(i, \mathbf{y}_{jab})} = \sum_{\mathbf{y} \setminus \mathbf{y}_{jab}} \mu_{(i,\mathbf{y})}$, where $\mathbf{y}_{jab}$ denotes the subconfiguration of $\mathbf{y}$ that matches the local function $j$ on pattern $a\mathbf{b}$. This allows a compact reformulation of an equivalent convex problem that can be solved efficiently as a compact quadratic program (Taskar et al., 2003). Unfortunately, this approach does not work readily in our case because the Lagrangian (15) does not permit a simple closed form expression of the dual.

Instead we have to follow a log-barrier approach to solving this problem (18). Since $\mathbf{\Delta w} \geq \gamma \boldsymbol{\delta} - S\boldsymbol{\epsilon}$ is too large to handle explicitly, an approach must be developed for handling the constraints implicitly. Unfortunately, a direct factorization approach is not readily available for reducing the exponential sum in $\sum_{(i,\mathbf{y})} \log \left( \mathbf{\Delta}(i, \mathbf{y})\mathbf{w} - \gamma \delta_{(i,\mathbf{y})} + \epsilon_i \right)$. Nevertheless the constraint generation strategy of (Altun et al., 2003) can be usefully applied in this case.

To solve (14) in the multivariate case we implemented a cutting plane method, where initially only a small subset of constraints in $\mathbf{\Delta w} \geq \gamma \boldsymbol{\delta} - S\boldsymbol{\epsilon}$ were considered. Given a current set of constraints, a solution $(\mathbf{w}, \gamma, \boldsymbol{\epsilon})$ was computed using the barrier method outlined above. Then for each training example $(\mathbf{x}^i, \mathbf{y}^i)$ one new labeling $\mathbf{y}$ was generated to maximize the degree of constraint violation

$$\begin{aligned} &\max_{\mathbf{y}} \; \gamma \delta_{(i,\mathbf{y})} - \epsilon_i - \mathbf{\Delta}(i, \mathbf{y})\mathbf{w} \\ =\;& \max_{\mathbf{y}} \; \exp \left( \gamma \delta_{(i,\mathbf{y})} + \boldsymbol{\phi}(\mathbf{x}^i, \mathbf{y})^\top \mathbf{w} \right) \quad (20) \end{aligned}$$

This is in fact an inference problem that can be solved by conventional methods. For example, if $\mathbf{y}$ forms a Markov chain, then the optimal constraint can be generated by a Viterbi algorithm run on the probability model defined by (20).

Once the new constraints have been generated, they are added to the problem and the solution $(\mathbf{w}, \gamma, \boldsymbol{\epsilon})$ is recomputed using the barrier method. In our experiments we found this constraint generation scheme to be quite effective, requiring at most 10 to 50 generation iterations to be executed before solving the problem to within small tolerances.

## 9 Experimental results

We implemented this approach and tested it on both synthetic and real data using HMM models for classification, where the classification variables $\mathbf{y}$ play the role of the hidden state sequence, and the input variables $\mathbf{x}$ play the role of the observations. Generally we considered models of the form depicted in Figure 8, where each $y$ variable has multiple (disjoint) $x$-children. In our synthetic experiment, we sampled $(\mathbf{x}, \mathbf{y})$ from a sequence of length 5 (5 $y$ variables with 4 $x$-children each, for a total of 20 $x$ variables). We then used a generative model based on the same skewed parameters used in the synthetic single-variable experiments above; here with $\beta = 0.85$. We repeated the experiment 20 times to obtain the final results. Figure 9 shows that M$^2$BN again outperforms M$^3$N and MCL in controlled experiments where the correct Bayesian network structure is known.

We also conducted an experiment on a protein secondary structure database (Cuff & Barton, 1999). Here the goal is to predict the sequence of secondary structure labels given an observed amino acid sequence. Figure 8 shows the prediction model we used. Basically, the secondary structure tag $y_k$ for a location $k$ in the amino acid sequence is predicted based on a sliding window of 7 adjacent locations, as well as the neighboring secondary structure tags. We trained on a subset of the data and tested on 1000 remaining locations disjoint from the training data. The experiment was repeated 20 times. Figure 10 shows that M$^2$BN is competitive with M$^3$N and MCL on this data set.

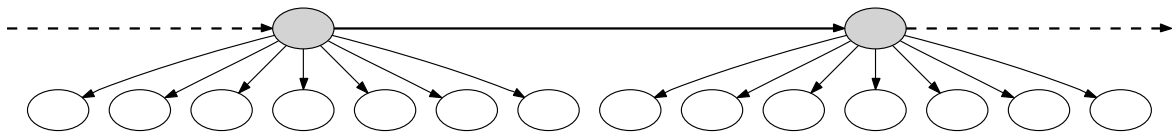

Figure 8: Structure of the protein secondary structure prediction model

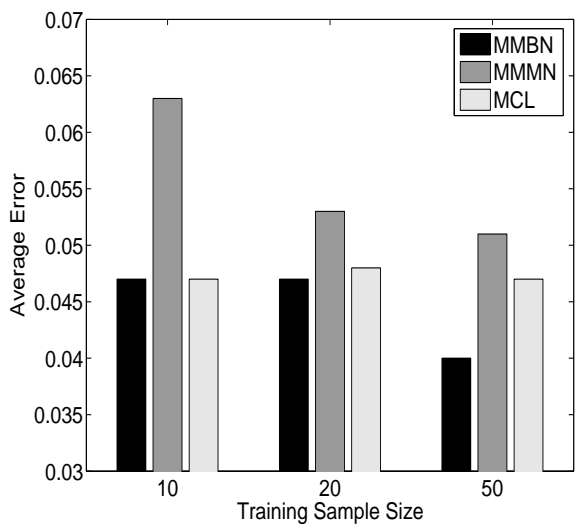

Figure 9: Average error results for $M^2BN$ and $M^3N$ on synthetic multi-variable networks.

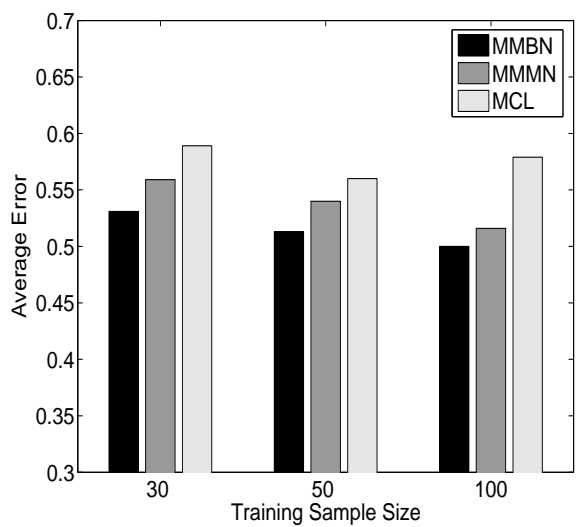

Figure 10: Average error results for $M^2BN$ and $M^3N$ on protein secondary structure prediction.

## 10 Conclusion

We have investigated what we feel is a very natural question; whether a Bayesian network representation can be combined with discriminative training based on the maximum margin criterion of SVMs. We have found that training Bayesian networks under the maximum margin criterion is a hard computational problem—harder than the standard quadratic program of SVM training. However, reasonable training algorithms can be devised which optimize the margin exactly in special cases, and provides a reasonable heuristic in general cases. Our preliminary experiments show that there might be an advantage to respecting the causal model constraints embodied by a Bayesian network, if indeed these constraints were present during the data generation. In this sense, maximum margin Bayesian networks offer a new way to add prior knowledge to SVMs.

The main directions for future research are to improve the training procedure and add kernels. As in (Taskar et al., 2003), it is possible to consider adding kernels to our local feature representation. The results of (Altun et al., 2004) show that it is possible to represent our local functions in terms of weighted combinations of training features. Therefore, one extension we are working on is to formulate hybrid classifiers with general kernelized potential functions between **y** and **x**, but standard conditional probability tables between classification variables to preserve their causal relationships.

### Acknowledgments

Research supported by the Alberta Ingenuity Centre for Machine Learning, NSERC, MITACS, CFI and the Canada Research Chairs program.

### References

Altun, Y., Smola, A., & Hofmann, T. (2004). Exponential families for conditional random fields. *Proceedings of the Conference on Uncertainty in Artificial Intelligence (UAI-04)*.

Altun, Y., Tsochantaridis, I., & Hofmann, T. (2003). Hidden Markov support vector machines. *Proceedings of the International Conference on Machine Learning (ICML-03)*.

Boyd, S., & Vandenberghe, L. (2004). *Convex optimization*. Cambridge U. Press.

Cheng, J., & Greiner, R. (1999). Comparing bayesian network classifiers. *Proceedings of the Conference on Uncertainty in Artificial Intelligence (UAI-99)*.

Collins, M. (2002). Discriminative training methods for hidden markov models: Theory and experiments with perceptron algorithms. *Proceedings of the Conference on Empirical Methods in Natural Language Processing (EMNLP-02)*.

Crammer, K., & Singer, Y. (2001). On the algorithmic interpretation of multiclass kernel-based vector machines. *Journal of Machine Learning Research*, *2*.

Cuff, J., & Barton, G. (1999). Evaluation and improvement of multiple sequence methods for protein secondary structure prediction. *Proteins: Structure, Function and Genetics*, *34*, 508–519. www.compbio.dundee.ac.uk/∼www-jpred/data.

Friedman, N., Geiger, D., & Goldszmidt, M. (1997). Bayesian network classifiers. *Machine Learning*, *29*, 131–163.

Lafferty, J., Liu, Y., & Zhu, X. (2004). Kernel conditional random fields: Representation, clique selection, and semi-supervised learning. *Proceedings of the International Conference on Machine Learning (ICML-04)*.

Lafferty, J., McCallum, A., & Pereira, F. (2001). Conditional random fields: Probabilistic models for segmenting and labeling sequence data. *Proceedings of the International Conference on Machine Learning (ICML-01)*.

Lanckriet, G., Cristianini, N., Bartlett, P., Ghaoui, L., & Jordan, M. (2004). Learning the kernel matrix with semidefinite programming. *Journal of Machine Learning Research*, *5*.

Ng, A., & Jordan, M. (2001). On discriminative vs. generative classifiers. *Advances in Neural Information Processing Systems 14 (NIPS-01)*.

Raina, R., Shen, Y., Ng, A., & McCallum, A. (2003). Classification with hybrid discriminative / generative models. *Advances in Neural Information Processing Systems 16 (NIPS-03)*.

Schoelkopf, B., & Smola, A. (2002). *Learning with kernels: Support vector machines, regularization, optimization, and beyond*. MIT Press.

Taskar, B., Guestrin, C., & Koller, D. (2003). Max-margin Markov networks. *Advances in Neural Information Processing Systems 16 (NIPS-03)*.

Taskar, B., Klein, D., Collins, M., Koller, D., & Manning, C. (2004). Max-margin parsing. *Proceedings of the Conference on Empirical Methods in Natural Language Processing (EMPNLP-04)*.

Tsochantaridis, I., Hofmann, T., Joachims, T., & Altun, Y. (2004). Support vector machine learning for interdependent and structured output spaces. *Proceedings of the International Conference on Machine Learning (ICML-04)*.

Vanderbei, R. (1996). *Linear programming: Foundations and extensions*. Kluwer.

Wettig, H., Grunwald, P., Roos, T., Myllymaki, P., & Tirri, H. (2002). *On supervised learning of Bayesian network parameters* (Technical Report HIIT 2002-1). Helsinki Inst. Info. Tech.

Wettig, H., Grunwald, P., Roos, T., Myllymaki, P., & Tirri, H. (2003). When discriminative learning of Bayesian network parameters is easy. *Proceedings of the International Joint Conference on Artificial Intelligence (IJCAI-03)*.